\documentclass[11pt,a4paper]{article}
\usepackage[hyperref]{naaclhlt2019}
\usepackage{times}
\usepackage{latexsym}
\usepackage{times}
\usepackage{helvet}
\usepackage{courier}
\usepackage{times}
\usepackage{latexsym}
\usepackage{graphicx}
\usepackage{amsmath}
\usepackage{graphicx}
\usepackage{array}
\usepackage{makecell}
\usepackage{fontawesome}

\usepackage{url}

\aclfinalcopy 

\title{Direct Network Transfer: Transfer Learning of Sentence Embeddings for Semantic Similarity}

\author{Li Zhang \\
  {} \\\And
  Steven R. Wilson \\
  University of Michigan \\
  {\tt \{zharry,steverw,mihalcea
\}@umich.edu} \\\And
  Rada Mihalcea \\
  {} \\}

\date{}

\begin{document}
\maketitle
\begin{abstract}
Sentence encoders, which produce sentence embeddings using neural networks, are typically evaluated by how well they transfer to downstream tasks. This includes semantic similarity, an important task in natural language understanding. Although there has been much work dedicated to building sentence encoders, the accompanying transfer learning techniques have received relatively little attention. In this paper, we propose a transfer learning setting specialized for semantic similarity, which we refer to as direct network transfer. Through experiments on several standard text similarity datasets, we show that applying direct network transfer to existing encoders can lead to state-of-the-art performance. Additionally, we compare several approaches to transfer sentence encoders to semantic similarity tasks, showing that the choice of transfer learning setting greatly affects the performance in many cases, and differs by encoder and dataset. 
\end{abstract}

\section{Introduction} \label{introduction}
	
In recent years, with the rise of neural networks, word embeddings or distributional word representations have become relatively mature and essential in many NLP tasks. On the other hand, creating sentence embeddings remains a difficult problem. Since the meaning of sentences depends not only on the individual words but also on the sequence of and interactions between words, there has been increasing interest in compositional models to embed sentences, or sentence encoders.

Sentence encoders are usually evaluated on downstream tasks such as sentiment analysis \cite{pang2004sentimental}, opinion mining \cite{hu2004mining}, and semantic similarity. 
Semantic similarity, or relating short texts in a semantic space -- be those phrases, sentences or short paragraphs -- is a task that requires systems to determine the degree of equivalence between the underlying semantics of the two texts.\footnote{In this paper, we use the word ``sentence'' loosely as the umbrella term ``short text'', unless otherwise noted.}
Although relatively easy for humans, this task remains one of the most difficult natural language understanding problems, receiving significant interest from the research community. For instance, from 2012 to 2017, the International Workshop on Semantic Evaluation (SemEval) has been holding a shared task called Semantic Textual Similarity (STS) \cite{Agirre12}, dedicated to tackling this problem, with around 100 team submissions each year (Table~\ref{semeval-submissions}).

\begin{table}[t!]
	\begin{center}
		\begin{tabular}{c|c|c}
			\hline \bf Year & \bf Task no. & \bf Submissions   \\ \hline
			2012 & \#6 & 89\\
			2013 & \#6 & 90\\
			2015 & \#2 & 74\\
			2016 & \#1 & 124\\
			2017 & \#1 & 85\\\hline
		\end{tabular}
	\end{center}
	\caption{\label{semeval-submissions} Count of valid submissions to and the numbering of each SemEval STS task from 2012 to 2017.}
\end{table}

While word embeddings have achieved near human-level performance on some word-level semantic similarity tasks \cite{Wieting15}, measuring the degree of equivalence in the underlying semantics of paired texts represents a challenging problem in NLP. There are two major methods that are typically used to compute the semantic similarity of two sentences. The first is to encode each sentence into a fixed-length vector \textit{independently} before computing the similarity between them. A wide range of approaches have been used to generate the sentence embeddings, such as unsupervised encoders \cite{Kiros15}, supervised encoders \cite{Conneau17} and translation-based approaches trained on multilingual corpora \cite{Hill14}. The second major approach is to \textit{jointly} take both sentences as input, using interactions between sentences (e.g., alignments, attention or other features like word overlap) to produce the similarity score \cite{tian2017ecnu}. While these joint approaches have been successful, one advantage of the independent method is that the produced embeddings can be used in downstream tasks where the representation of a single sentence is required. Since we study how sentence encoders transfer to semantic similarity tasks, we focus on the first approach. 

\begin{figure}[t!]
	\includegraphics[width=0.48\textwidth]{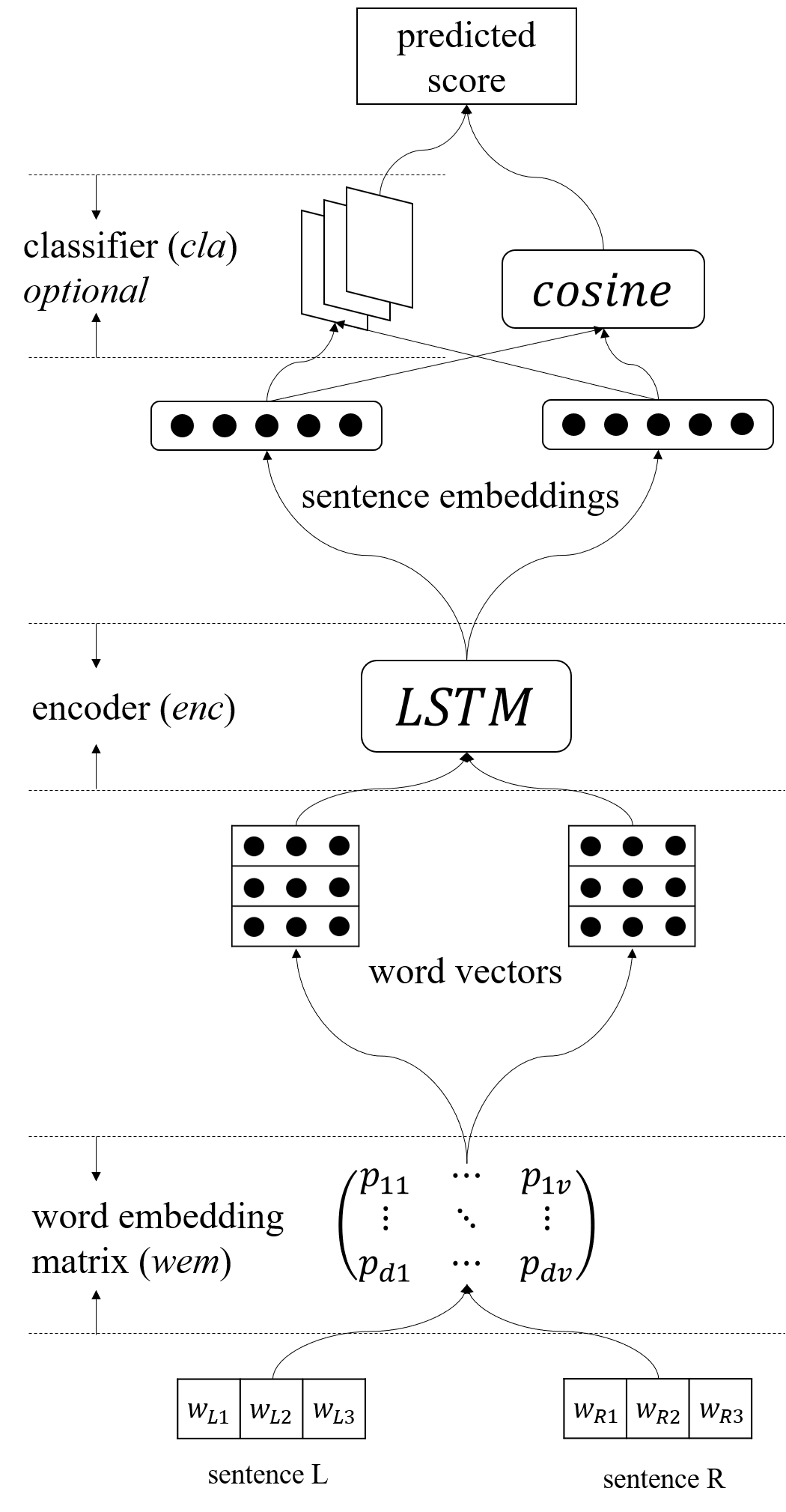}
	\caption{\label{encoder-arch-fig} Overview of generic sentence encoder architecture with \textit{word embedding matrix}, \textit{encoder} and optionally \textit{classifier}.}
\end{figure}

A typical compositional sentence encoder consists of two components: a \textit{word embedding matrix}, which is a mapping between words and fixed-length vectors, and an \textit{encoder}, which takes a sequence of word embeddings and outputs a single sentence embedding.\footnote{We use italics to disambiguate the components of a sentence encoder.} During initial training, the \textit{word embedding matrix} is usually initialized with pre-trained word embeddings, while the \textit{encoder} is learned via a large corpus. The sentence embeddings produced by the \textit{encoder} can then be used to evaluate on semantic similarity tasks. When transferring to these tasks, a \textit{classifier} is often added on top of the encoder to produce the predicted similarity score. Each of the three components has its own set of network parameters, denoted as \textit{wem}, \textit{enc} and \textit{cla} (Figure~\ref{encoder-arch-fig}).

Transfer learning for semantic similarity tasks is typically implemented by fixing the weights of a pre-trained network and adding an additional \textit{classifier} on top, which is trained on the downstream dataset. However, we propose direct network transfer, a new transfer setting in which a model is directly trained to optimize the measure of similarity between sentence embeddings (changing \textit{cla} to \textit{cosine} in Figure~\ref{encoder-arch-fig}) rather than treating similarity prediction as a regression task. In this setting, the model parameters (\textit{wem} and \textit{enc}) are updated through back-propagation, effectively tuning the sentence encoder itself rather than learning a transformation of a fixed, pre-trained representation. By experimenting with various combinations of sentence encoders and semantic similarity datasets, we show that, in many cases, direct network transfer is superior to learning an additional output layer, i.e, \textit{cla}.

This paper makes two key contributions. First, we propose a novel transfer learning setting specifically designed for semantic similarity tasks called direct network transfer, with which we achieve state-of-the-art results on the STS Benchmark \cite{Cer17} and the Human Activity Phrase dataset \cite{Wilson17}. After experimenting with many combinations of sentence encoders and semantic similarity datasets, we find that direct network transfer outperforms existing transfer settings in most cases. Second, we compare a range of transfer learning settings for semantic similarity using some of the best sentence encoders to date across an array of datasets, and we explain the patterns by which the performances vary. We also explore the details of the transfer learning architecture by proposing transfer related hyperparameters and recommendations on how to choose them.
	
\section{Related Work} \label{related-work}

Following Pan and Yang \shortcite{pan2010survey}, we denote the source dataset that a sentence encoder has trained on as $S$ (usually large), and the semantic similarity target dataset as $T$ (usually small). There are typically three settings of transferring a sentence encoder to a semantic similarity task.

\subsection{Unsupervised evaluation} The model is only trained on $S$ and then evaluated on $T$. During evaluation, some distance metric is calculated between the embeddings of two sentences as the predicted score. In this setting, \textit{wem} and \textit{enc} are frozen, meaning that they do not receive gradients and are not updated, and \textit{cla} does not exist. Technically, no transfer \textit{learning} is applied.

\subsection{Feature transfer} The model is first trained on $S$, learning \textit{wem} and \textit{enc} in the process. When transferring to $T$, a \textit{classifier} with randomly initialized weights is trained to make predictions using the sentence embeddings produced by the \textit{encoder} as input features. This is equivalent to using a new model whose \textit{wem} and \textit{enc} are initialized as learned in $S$ and whose \textit{cla} is initialized randomly. In this setting, \textit{wem} and \textit{enc} are frozen and only \textit{cla} is updated while training on $T$.

\subsection{Network transfer} This setting is also commonly called fine-tuning \cite{razavian2014cnn}. Like feature transfer, the model is trained on both $S$ and $T$ and evaluated on $T$, and a \textit{classifier} is added on top to produce the predicted score. However, while training on $T$, in addition to learning the \textit{cla} parameters, either \textit{enc} or both \textit{wem} and \textit{enc} are updated, while the other parameters are frozen.

Although most sentence encoders have been built with the goal of transferring to downstream tasks, the methodology of transfer learning itself is inconsistent \cite{mou2016transferable}. There has not been a consensus on which transfer learning setting to use with sentence encoders for semantic similarity tasks (Table~\ref{transfer-setting-table}). The two semantic similarity datasets popularly used for sentence embedding evaluation are SICK Relatedness and SemEval STS. Among sentence encoders that evaluate on these two tasks within three years, SIF on GloVE \cite{arora2016simple} and ParaNMT \cite{Wieting17pushing} use unsupervised evaluation on STS but use feature transfer on SICK; InferSent \cite{Conneau17} use feature transfer on both; GRAN \cite{Wieting17revisit} and Sent2vec \cite{pagliardini2017unsupervised} use unsupervised evaluation on both; Paragram \cite{Wieting15a} is the only work, to our knowlege, that experiments on both feature transfer and network transfer for both datasets; Cer et al. \shortcite{cer2018universal} use unsupervised evaluation while Logeswaran and Lee \shortcite{logeswaran2018efficient} use feature transfer on SICK, but they do not evaluate on STS.

\begin{table}[t!]
	\begin{center}
	\small
		\begin{tabular}{l|l|l}
			\hline  & SICK & STS \\ \hline
			\cite{cer2018universal} & UE & -\\
			\cite{logeswaran2018efficient} & FT & -\\
			\cite{pagliardini2017unsupervised} & UE & UE\\
			\cite{Wieting17revisit} & UE & UE\\
			\cite{Conneau17} & NT & NT\\
			\cite{Wieting17pushing} & FT & UE\\
			\cite{arora2016simple} & FT & UE\\
			\cite{Wieting15a} & FT, NT & UE\\
			\cite{Kiros15} & - & -\\
			\hline
		\end{tabular}
	\end{center}
	\caption{Transfer learning settings used by sentence encoders on semantic similarity tasks in recent years, ranked by time of publication. UE: unsupervised evaluation; FT: feature transfer; NT: network transfer.}
	\label{transfer-setting-table}
\end{table}
	
\section{Direct Network Transfer} \label{DNT}

We propose a novel transfer learning setting called \textit{direct network transfer} specialized for semantic similarity tasks, in which the cosine similarity of sentence embedding pairs is directly used in the loss function during transfer learning. This is not to be confused with a similar approach known as \textit{fine-tuning} or \textit{network transfer}, in which embeddings are inputs to a logistic regression model (discussed in more detail later).

\subsection{Motivation}

Transferring to semantic similarity is unlike transferring to other downstream tasks in that the sentence embeddings already carry some information about similarity geometrically, expressed by the distance between vectors. For this reason, it is not necessary to have logistic regression layers to transform the embeddings. Instead, a fixed distance metric between the two embedding vectors can be used to produce a similarity score. Not only simpler, direct network transfer might also allow sentence encoders to optimize distance between embeddings - not just the embeddings themselves - leading to more effective transfer. 

\subsection{Definition} \label{DNT-definition}

Following the notation introduced in the Related Work section, the model is trained on both $S$ and $T$ and evaluated on $T$. Cosine similarity is calculated between the embeddings of two sentences as the prediction, meaning there is no \textit{classifier} as in network transfer or feature transfer. While training on $T$, some or all of \textit{wem} and \textit{enc} are updated while the rest are frozen. 

The optimization goal is that the cosine similarity of sentence embeddings should be close to the normalized annotated score. We normalize the annotated scores in each dataset to $[0,1]$, with more discussion on normalization below. With sentence embeddings $h_L$ and $h_R$, the loss is calculated as: 
\begin{equation} \label{dnt-loss}
   \frac{1}{m} \sum_{k=1}^{m}(\text{cosine}(h_L^{(k)},h_R^{(k)})-y')^2
\end{equation}
where $m$ is the number of pairs in a batch, $k$ indicates the $k$-th sentence pair and $y'$ is the normalized annotated score.  
	
\section{Setup}

We evaluate various sentence encoders under various transfer settings on a collection of semantic similarity target datasets.

\subsection{Datasets}

Each semantic similarity dataset typically consists of a set of sentence pairs, with an annotated similarity score for each pair. The task is defined as follows: a model takes a pair of sentences as input and predicts the similarity score. Then, a correlation coefficient (Pearson's $r$ or Spearman's $\rho$) is calculated between the predicted scores and the annotated scores for all pairs. The goal is to maximize this correlation coefficient. 

We consider the following datasets, which are popular in sentence encoder literature.\\
\textbf{STS Benchmark} \cite{Cer17}: a selection of the English datasets used in the STS tasks organized by SemEval 2012 - 2017, designed to be a standard benchmark for the evaluation of text similarity systems\footnote{http://ixa2.si.ehu.es/stswiki/index.php/STSbenchmark}. The annotated scores range from $0$ to $5$, and results are reported in terms of Pearson's $r$. There are 5749 pairs in the train split, 1500 in the development split and 1379 in the test split. \\
\textbf{SICK relatedness} \cite{Marelli14a}: the Sentences Involving Compositional Knowledge benchmark, which includes a large number of sentence pairs that are rich in the lexical, syntactic and semantic phenomena specifically designed to test the ability of systems to model semantic compositionality. Each pair is annotated for semantic relatedness with scores ranging from $1$ to $5$, and Pearson's $r$ is used for evaluation. There are 4439 pairs in the train split, 495 in the development split and 4906 in the test split. \\
\textbf{STS 2012} \cite{Agirre12}: the first Semantic Textual Similarity (STS) shared task to examine the degree of semantic equivalence between two sentences in Semantic Evaluation (SemEval). Compared to the other years of SemEval STS tasks, this particular one has both training and testing data while the tasks from other years only provide testing data, making STS 2012 ideal for transfer learning evaluation. The annotated scores range from $0$ to $5$, and Pearson's $r$ is used. There are 2234 pairs in the original train split in which we use 234 as the development split and the rest as the train split, and 1959 in the test split. 

While the datasets above focus on general textual similarity, we also include some focusing on fine-grained semantic similarity.\\
\textbf{Human Activity Phrase} \cite{Wilson17}: a collection of pairs of phrases describing human activities, annotated in four different dimensions, treated as separate tasks: \\
Similarity (SIM): The degree to which the two activity phrases describe the same thing, semantic similarity in a strict sense. Example of high similarity phrases: \textit{to watch a film} and \textit{to see a movie}. \\
Relatedness (REL): The degree to which the activities are related to one another, a general semantic association between two phrases. Example of strongly related phrases: \textit{to give a gift} and \textit{to receive a present}. \\
Motivational Alignment (MA): The degree to which the activities are (typically) done with similar motivations. Example of phrases with potentially similar motivations: \textit{to eat dinner with family members} and \textit{to visit relatives}.\\
Perceived Actor Congruence (PAC): The degree to which the activities are expected to be done by the same type of person. An example of a pair with a high PAC score: \textit{to pack a suitcase} and \textit{to travel to another state}.\\
The annotated scores range from $0$ to $4$ for SIM, REL and MA (unipolar scales), and $-2$ to $2$ for PAC (bipolar scale), and Spearman's $\rho$ is used for evaluation. There are 1000 pairs in the dataset. Since this test set is already small, we collected 1373 additional annotated pairs of human activity phrases in the same format as this dataset, randomly choosing 1000 for training and 373 for development.\footnote{We make this set of training pairs available at: http://anonymous.edu/.} We then treat the original 1000 pairs as a held-out test set so that our results are directly comparable with those previously reported. \\
\textbf{Short Answer Grading Dataset} \cite{mohler2011learning}: a collection of student and instructor answers to questions on assignments and examinations in a Data Structures course. The student answers were independently graded by two human judges based on their consistency with the instructor answers. For this dataset, a training instance consists a student/instructor answer pair and a graded score. The annotated scores range from $0$ to $5$, and Pearson's $r$ is used for evaluation. There are 2273 pairs in the dataset which come from 10 assignments and 2 exams. We randomly sample 1460 pairs from the assignments as the train split, use the rest (500 pairs) as the development split, and use the exams (552 pairs) as the test split. This is not directly comparable to the results reported in the paper of this dataset, as they used 12-fold cross-validation which became too computationally expensive to fully replicate when training deep neural networks. 
	
\subsection{Models}
	
We experiment on the following pre-trained sentence encoders, which have recently achieved state-of-the-art results on various downstream tasks, including semantic similarity.\\
\textbf{Infersent} \cite{Conneau17}: a bi-directional LSTM with max pooling trained on the Stanford Natural Language Inference (SNLI) dataset \cite{Bowman15} and Multi-Genre Natural Language Inference corpus \cite{williams2017broad}.\\
\textbf{Gated Recurrent Averaging Network}\ (GRAN) \cite{Wieting17revisit}: a paraphrastic compositional model that combines LSTM and averaging word embeddings, trained on sentence pairs obtained by aligning Simple English to standard English Wikipedia (Simple-Wiki dataset) \cite{Coster11}. \\
\textbf{BiLTSM-Avg} \cite{Wieting17pushing,Wieting17revisit}: a bi-directional LSTM model that averages all hidden vectors to generate the sentence embedding which has a large dimension of 4096, trained on the back-translated Czeng1.6 corpus \cite{Bojar16} (PARANMT-50M).
	
\subsection{Other Transfer Settings}

As discussed in Related Work, there are typically three transfer settings given a sentence encoder and a target dataset. In addition to our new direct network transfer setting described before, we also compare the three other settings: unsupervised evaluation, feature transfer, and network transfer. The implementation details are as follows. \\
\textbf{Unsupervised evaluation}: the pre-trained sentence encoder is directly applied to evaluate on the semantic similarity datasets without any further training, using cosine similarity to produce similarity scores.\\
\textbf{Feature transfer}: we use the code from SentEval \footnote{https://github.com/facebookresearch/SentEval} which follows the methods of Tai et al. \shortcite{Tai15}. The sentence embeddings produced by the sentence encoder are used as fixed input features to a \textit{classifier} consisting of a dense layer and a softmax layer. Given a score for a sentence pair in the range $[1,K]$, where $K$ is an integer, with sentence embeddings $h_L$ and $h_R$, and model parameters $\theta$, we compute:
	
\begin{align*}
    \begin{split}
     h_\times = h_L \odot h_R,\ h_+ = |h_L - h_R| , \\
     h_s = \sigma (W^{(\times)}h_\times + W^{(+)}h_+ + b^{(h)}), \\
     \hat{p}_\theta = \text{softmax}(W^{(p)}h_s+b^{(p)}), \\
     \hat{y} = r^T \hat{p}_\theta ,
    \end{split}
\end{align*}
where $W^{(\times)},W^{(+)},b^{(h)},W^{(p)},b^{(p)}$ are parameters of the layers, $r^T=[1,2\dots K]$, and $\hat{y}$ is the predicted score. To encode the annotated score as a distribution, a sparse target distribution $p$ is defined such that it satisfies $y=r^T p$ where $y$ is the annotated score:
\[
    p_i= 
\begin{cases}
    y-\lfloor y \rfloor, & i=\lfloor y \rfloor+1\\
    \lfloor y \rfloor-y+1, & i=\lfloor y \rfloor\\
    0, & \text{otherwise}
\end{cases}
\]
for $1\leq i \leq K$. The mean squared error loss (used by SentEval) or the Kullback-Leibler divergence loss (used by Tai et al. \shortcite{Tai15}) is calculated between $p$ and $\hat{p}_\theta$:
\begin{equation} \label{loss}
	\frac{1}{m} \sum_{k=1}^{m} \text{Loss}(p^{(k)}\|\hat{p}_\theta^{(k)})
\end{equation}
where $m$ is the number of pairs in a batch and $k$ indicates the $k$-th sentence pair. We compare the performances using both loss functions.\\
\textbf{Network transfer}: instead of using sentence embeddings as fixed input features, the \textit{word embedding matrix} and \textit{encoder} which produce the sentence embeddings, alongside the \textit{classifier}, constitute an end-to-end model. In this model, all three sets of parameters (\textit{wem}, \textit{enc}, \textit{cla}) can be updated (receiving gradients) or frozen (not receiving gradients). We compare two scenarios: 1. all three are updated, and 2. only \textit{wem} is frozen. We do not report the results of only freezing \textit{enc} because they are higher layers of the neural network and freezing them without also freezing \textit{wem} leads to undesirable performance.

A comparison of the transfer settings discussed is in Table~\ref{transfer-details}. 

\begin{table}[t!]
	\begin{center}
	\small
		\begin{tabular}{l|l|l}
			\hline  & \faLock\ params. & \faUnlock\ params. \\ \hline
			Unsupervised evaluation & \textit{wem}, \textit{enc} & -\\
			Feature transfer & \textit{wem}, \textit{enc} & \textit{cla}\\
			Network transfer \faLock & - & \textit{wem}, \textit{enc}, \textit{cla}\\
			Network transfer \faUnlock & \textit{wem} & \textit{enc}, \textit{cla}\\\hline
			This work &  & \\\hline
			Direct network transfer \faLock & - & \textit{wem}, \textit{enc}\\
			Direct network transfer \faUnlock & \textit{wem} & \textit{enc}\\
			\hline
		\end{tabular}
	\end{center}
	\caption{Transfer learning settings with the corresponding parameter set(s) being frozen (lock icon) or updated (unlock icon). }
	\label{transfer-details}
\end{table}

\section{Experiments}

To understand how the choice of transfer setting plays a significant role in transfer learning of sentence encoders on semantic similarity tasks and how direct network transfer performs compared to other settings, we experiment on each combination of dataset, model and transfer setting outlined above. 

\begin{table*}[t!]
	\begin{center}
		\begin{tabular}{l|l|l|l|l|l|l|l|l}
			\hline \bf Datasets & \bf STS Bench. & \bf SICK  & \bf STS 12  & \bf SIM  & \bf REL  & \bf MA  & \bf PAC  & \bf SAG \\ \hline
			BiLSTM-Avg [UE] & .791/.783 & .735 & .803 & .649 & .639 & .603 & .469 & .450 \\
			BiLSTM-Avg [FT] MSE & .779/.746 & .860 & \underline{\textbf{.867}}$\dagger$ & .534 & .514 & .474 & .412 & .761 \\
			BiLSTM-Avg [FT] KL & .797/.779 & .861 & .864 & .518 & .509 & .461 & .400 & .774 \\
			BiLSTM-Avg [NT] MSE \faLock & .836/.810 & .864 & .860 & .576 & .575 & .529 & .456 & .761 \\
			BiLSTM-Avg [NT] MSE \faUnlock & .833/.809 & .864 & .861 & .571 & .571 & .526 & .453 & .806 \\
			BiLSTM-Avg [NT] KL \faLock & .840/.806 & \textbf{.866} & .854 & .559 & .558 & .515 & .459 & .801 \\
			BiLSTM-Avg [NT] KL \faUnlock & .837/.802 & .864 & .845 & .556 & .529 & .512 & .449 & .813 \\
			BiLSTM-Avg [DNT] \faLock & \underline{\textbf{.852/.824}}$\dagger$ & .856 & .861 & \textbf{.699} & \textbf{.688} & \textbf{.660} & \textbf{.470} & .816 \\
			BiLSTM-Avg [DNT] \faUnlock & \underline{\textbf{.851/.824}}$\dagger$ & .859 & .861 & .691 & .680 & .646 & .462 & \textbf{.834} \\\hline
			GRAN [UE] & .688/.583 & .703 & .560 & .644 & .642 & .596 & \textbf{.444} & .323 \\
			GRAN [FT] MSE & .759/.693 & .792 & .651 & .561 & .576 & .526 & .392 & .504 \\
			GRAN [FT] KL & \textbf{.771/.701} & .790 & .649 & .556 & .577 & .525 & .398 & .649 \\
			GRAN [NT] MSE \faLock & .710/.648 & \textbf{.857} & \textbf{.734} & .575 & .567 & .523 & .375 & .742 \\
			GRAN [NT] MSE \faUnlock & .720/.653 & .855 & .726 & .578 & .560 & .510 & .385 & .736 \\
			GRAN [NT] KL \faLock & .717/.643 & \textbf{.857} & .731 & .558 & .574 & .530 & .401 & .802 \\
			GRAN [NT] KL \faUnlock & .717/.644 & .853 & .718 & .541 & .537 & .442 & .415 & .791 \\
			GRAN [DNT] \faLock & .749/.644 & \textbf{.857} & .670 & \textbf{.668} & .663 & \textbf{.624} & .407 & .792 \\
			GRAN [DNT] \faUnlock & .745/.641 & \textbf{.857} & .663 & \textbf{.668} & \textbf{.666} & \textbf{.623} & .413 & \textbf{.807} \\\hline
			InferSent [UE] & .782/.738 & .748 & .607 & \underline{\textbf{.701}}$\dagger$ & .686 & .652 & .525 & .209 \\
			InferSent [FT] MSE & .809/.757 & \underline{\textbf{.884}}$\dagger$ & \textbf{.792} & .655 & .644 & .608 & .432 & .738 \\
			InferSent [FT] KL & \textbf{.831/.783} & .882 & .788 & .688 & .680 & .642 & .510 & .735 \\
			InferSent [NT] MSE \faLock & .783/.744 & .859 & .777 & .699 & .692 & .672 & .537 & .792 \\
			InferSent [NT] KL \faLock & .812/.763 & .867 & \textbf{.791} & .679 & .668 & .634 & .484 & .783 \\
			InferSent [DNT] \faLock & .802/.740 & .854 & .742 & \underline{\textbf{.702}}$\dagger$ & \underline{\textbf{.722}}$\dagger$ & \underline{\textbf{.691}}$\dagger$ & \underline{\textbf{.572}}$\dagger$ & \underline{\textbf{.838}}$\dagger$ \\\hline
		\end{tabular}
	\end{center}
	\caption{\label{results} The performance of transfer settings for three models across all datasets. Spearman's $\rho$ is reported for Human Activity Phrase dataset including the four dimensions SIM, REL, MA and PAC, and Pearson's $r$ for the rest, in accordance with the specification of the dataset to allow for direct comparison with previous results. The lock icon indicates freezing the \textit{word embedding matrix} weights (\textit{wem}), and the unlock icon indicates updating them. Note that \textit{wem} of InferSent must be frozen due to its implementation constraints. For each dataset, the best transfer result per-model is listed in bold font, the best overall result is underlined, and the state-of-the-art result is marked by a dagger.}
\end{table*}

\subsection{Experimental Details}
In each experiment, we use Adam \cite{kingma2014adam} as optimizer and tune the batch size over $\{32,64\}$, the learning rate over $\{0.1, 0.01, 0.001, 0.0001\}$ and the number of epochs over $\{10,30,50\}$. For each dataset in the rest of this paper, we tune these hyperparameters on the development set. When the transfer setting is feature transfer, network transfer or direct network transfer, we experiment with both MSE loss and KL Divergence loss and both freezing and updating \textit{word embedding matrix} weights. However, the architecture of InferSent uses a fixed \textit{word embedding matrix}, meaning that it has to be frozen. We use early stopping as regularization. All hyperparameters not mentioned maintain their values from the original code. 

\begin{table*}[t!]
	\begin{center}
		\begin{tabular}{l|l|l|l|l|l|l|l|l}
			\hline \bf Datasets & \bf STS Bench. & \bf SICK  & \bf STS 12  & \bf SIM  & \bf REL  & \bf MA  & \bf PAC  & \bf SAG \\ \hline
			BiLSTM-Avg [DNT] \faLock & \makecell{.852/.824 \\ .828/.800} & \makecell{.856 \\ .789} & \makecell{.861 \\ .824} & \makecell{.699 \\ .626} & \makecell{.688 \\ .622} & \makecell{.660 \\ .591} & \makecell{.470 \\ \textbf{.541}} & \makecell{.816 \\ \textbf{.823}} \\ \hline
			BiLSTM-Avg [DNT] \faUnlock & \makecell{.851/.824 \\ .827/.801} & \makecell{.859 \\ .793} & \makecell{.861 \\ .830} & \makecell{.691 \\ .625} & \makecell{.680 \\ .621} & \makecell{.646 \\ .587} & \makecell{.462 \\ \textbf{544}} & \makecell{.834 \\ .815} \\\hline
			GRAN [DNT] \faLock & \makecell{.749/.644 \\ .711/.604} & \makecell{.857 \\ .612} & \makecell{.670 \\ .544} & \makecell{.668 \\ .608} & \makecell{.663 \\ .607} & \makecell{.624 \\ .575} & \makecell{.407 \\ \textbf{.495}} & \makecell{.792 \\ .785} \\ \hline
			GRAN [DNT] \faUnlock & \makecell{.745/.641 \\ .720/.617} & \makecell{.857 \\ .617} & \makecell{.663 \\ .582} & \makecell{.668 \\ .601} & \makecell{.666 \\ .610} & \makecell{.623 \\ .596} & \makecell{.413 \\ \textbf{.488}} & \makecell{.807 \\ .758} \\\hline
			InferSent [DNT] \faLock & \makecell{.802/.740 \\ .556/.508} & \makecell{.854 \\ .653} & \makecell{.742 \\ .546} & \makecell{.702 \\ .537} & \makecell{.722 \\ .532} & \makecell{.691 \\ .628} & \makecell{.572 \\ .488} & \makecell{.838 \\ .798} \\\hline
		\end{tabular}
	\end{center}
	\caption{\label{norm-results} The effect of normalizing annotated scores to $[0,1]$ (upper side of each block) and to $[-1,1]$ (lower side). Performances in cases where normalizing to $[-1,1]$ performs better are in bold font. }
\end{table*}

\subsection{Comparison of Transfer Settings}

The results are shown in Table~\ref{results}. First and foremost, out of 24 combinations of sentence encoders and semantic similarity datasets, direct network transfer outperforms all other transfer settings in 16 cases, with a correlation coefficient difference smaller than $.002$ being counted as a tie. In 6 out of the 8 datasets, the best results overall are achieved with direct network transfer. State-of-the-art results are obtained when using direct network transfer with BiLSTM-Avg on STS Benchmark and with InferSent on all dimensions of Human Activity Phrase dataset. For general semantic similarity datasets including STS Benchmark, SICK and STS 12, direct network transfer performs on par with other settings in most combinations. For fine-grained datasets including the Human Activity Phrase dataset and Short Answer Grading dataset, direct network transfer outperforms all other settings in all but one case. This is valuable because being able to adapt to fine-grained and more difficult tasks shows that the model is able to capture semantic relationships that are more delicate than general similarity.

Among the other transfer settings, unsupervised evaluation only produces the best result in 2 out of 24 combinations. This shows that in most cases, not training on the target dataset during transfer learning is undesirable, more so if the target domain is greatly different from the source domain. For example, with InferSent on Short Answer Grading, the Pearson's $r$ increases by at least $.529$ when using any transfer setting that makes use of target training data. Due to the usually small size of target data, training on them is likely fast and fruitful. Therefore, unsupervised evaluation should only be used as a baseline when selecting transfer settings, unless there is no training data available in the target domain.

Feature transfer is a versatile setting that can be used with almost any downstream task. In the specific case of semantic similarity, however, our results indicate that it is often not the most effective setting. When transferring both BiLSTM-Avg and GRAN to the Human Activities dataset, feature transfer even underperforms unsupervised evaluation, which does not make use of target training data, in 7 out of 8 cases. However, feature transfer demonstrates better results when used with InferSent, with top performances on STS Benchmark, SICK and STS 12. One possible explanation is that during the original training (before transfer) of InferSent, the embeddings produced by the \textit{encoder} is already used as input features to the \textit{classifier} on top, as the two are jointly trained in an end-to-end fashion. Hence, the embeddings produced by InferSent is more suitable for use as inputs for a classifier than those produced by BiLSTM-Avg or GRAN, which do not employ a \textit{classifier} layer during their original training. However, even with InferSent, feature transfer is outperformed by direct network transfer in Human Activity Phrase dataset and Short Answer Grading dataset. 

Although network transfer is similar to direct network transfer, it is slower to train due to the extra parameters used to learn the \textit{classifier}. Though it does lead to good performance on STS Benchmark, SICK and STS 12, it is not as competitive when it comes to fine-grained semantic similarity datasets such as Human Activity Phrase dataset and Short Answer Grading dataset. 

\subsection{Annotated Score Normalization} \label{normalization}

Direct network transfer requires normalization of the annotated scores ($y'$ in Equation~\ref{dnt-loss}), since we want to optimize the cosine similarity between embeddings to be close to that value. Though cosine similarity values fall in the range $[-1,1]$, we observe that the embedding vectors are mostly non-negative, a trend that has also been noted for word embeddings \cite{Mimno17}. As a result, in Table~\ref{results} we normalize the annotated scores in each datasets to $[0,1]$. However, we also compare the effect of normalizing to $[-1,1]$.

As shown in Table~\ref{norm-results}, normalizing to $[0,1]$ is almost always superior except when the task is perceived actor congruence (PAC) of the Human Activity Phrase dataset, where normalizing to $[-1,1]$ is preferred for most combinations. One possible explanation is that in this dataset, while a similarity (SIM) score (in the range of $[0,4]$) of $0$ suggests that two activities are semantically unrelated, a PAC score (in the range of $[-2,2]$) of $-2$ suggests that two activities are very unlikely to be done often by the same type of person (e.g., ``have dinner with friends'' and ``eat dinner by oneself''), and so these pairs may actually exhibit a degree of semantic relatedness. If the score is normalized to $[0,1]$, $-2$ is transformed to $0$ meaning the two embedding vectors are pushed to be orthogonal, which is not desired. However, this choice is highly specific to datasets and practically should be tuned as a hyperparameter. 

\subsection{Effect of Freezing Lower Layers}

When transferring weights between neural networks, it is standard practice to sometimes freeze the lower layers and only update the higher layers to avoid both overfitting and catastrophic forgetting \cite{kirkpatrick2017overcoming}. This is possible with network transfer and direct network transfer, as the \textit{encoder} is being updated via back-propagation. For each combination using these two settings, we consider both freezing and updating the \textit{word embedding matrix}. 

As shown in Table~\ref{results}, there is no clear pattern on whether freezing or updating \textit{wem} is better. In most combinations, the differences between the two options are practically insignificant. Though freezing \textit{wem} theoretically results in fewer parameters to update and faster training, practically we found the difference in training time to be negligible, hence the choice between the two options should be tuned as a hyperparameter. 

\subsection{Effect of Loss Functions}

Unlike direct network transfer, which outputs a prediction as a real-valued cosine similarity, feature transfer and network transfer output a probability distribution over all score ranges, and thus there is a choice for loss function between MSE loss and KL-divergence loss (Equation~\ref{loss}). 

As shown in Table~\ref{results}, there is no clear pattern on whether MSE loss or KL-divergence loss is better. For InferSent, the difference between using the two losses are more significant than the other two models, especially on the Human Activity Phrase dataset. Practically, the choice between the two losses should be tuned as a hyperparameter. 

\subsection{Effect of Domains}

It has been standard practice to use feature transfer when the source and target domains are similar, and use network transfer when the source and target domains are dissimilar\footnote{http://cs231n.github.io/transfer-learning/}. Such pattern can be seen when transferring InferSent to SICK using network transfer, where the source and target domains are both captions and many of the pairs exhibit textual entailment or contradiction relationships \cite{Wieting17pushing}. However, all three sentence encoders that we consider are trained on large corpora whose domains are hard to classify, and the pattern is unclear in other combinations.

\section{Conclusion and Future Work}

We choose 3 sentence encoders, 8 semantic similarity tasks and 29 different transfer learning settings and evaluate on the performance on every combination. We propose a novel setting, direct network transfer, which achieve state-of-the-art on 5 tasks and outperforms all other settings in the majority of combinations. In addition, we systematically explore detailed changes that can be made to the transfer settings and discuss how they affect performance. 

Despite the success of direct network transfer, this work only compares models based on LSTM, which is a typical choice of architecture for sentence encoders. It would be worth exploring if direct network transfer also performs well when transferring non-LSTM based models to new datasets and domains. Furthermore, this work only considers semantic similarity tasks. Future work should explore transferring to other downstream tasks such as sentiment prediction or text retrieval.

\bibliography{naaclhlt2019}
\bibliographystyle{acl_natbib}

\end{document}